# AI-Enabled Waste Classification as a Data-Driven Decision Support Tool for Circular Economy and Urban Sustainability

1st Julius Sechang Mboli
*Centre of Excellence for Data Science, Artificial Intelligence and Modelling (DAIM)*
*University of Hull*
Hull, United Kingdom
mboli4god@gmail.com,
0000-0003-1708-3052

2nd Omolara Aderonke Ogungbemi
*Centre of Excellence for Data Science, Artificial Intelligence and Modelling (DAIM)*
*University of Hull*
Hull, United Kingdom

*Abstract*— Efficient waste sorting is crucial for enabling circular-economy practices and resource recovery in smart cities. This paper evaluates both traditional machine-learning (Random Forest, SVM, AdaBoost) and deep-learning techniques including custom CNNs, VGG16, ResNet50, and three transfer-learning models (DenseNet121, EfficientNetB0, InceptionV3) for binary classification of 25 077 waste images (80/20 train/test split, augmented and resized to 150×150 px). The paper assesses the impact of Principal Component Analysis for dimensionality reduction on traditional models. DenseNet121 achieved the highest accuracy (91 %) and ROC-AUC (0.98), outperforming the best traditional classifier by 20 pp. Principal Component Analysis (PCA) showed negligible benefit for classical methods, whereas transfer learning substantially improved performance under limited-data conditions. Finally, we outline how these models integrate into a real-time Data-Driven Decision Support System for automated waste sorting, highlighting potential reductions in landfill use and lifecycle environmental impacts.)

## I. Introduction

Efficient waste management lies at the heart of sustainable urban development. Rapid urbanisation, exponential proliferation of mobile devices, internet ubiquity and growing consumption patterns have led to an unprecedented increase in municipal solid waste, including e-waste thereby placing severe strain on landfills and exacerbating environmental degradation [1]. The circular-economy paradigm, whereby materials are retained in use for as long as possible through strategies of reuse, repair, remanufacture, recycling, and cascading offers a compelling framework to decouple economic growth from resource depletion [2]. However, realising a truly circular model requires accurate, automated sorting of waste streams to maximise recovery rates and minimise contamination.

Recent advances in computer vision and machine learning have driven significant improvements in automated image-based classification tasks. Convolutional neural networks (CNNs) and transfer-learning approaches, in particular, have demonstrated superior performance over traditional feature-engineering methods across diverse domains. Yet, their application to real-world waste classification remains underexplored, with prior studies often limited by small datasets, simplistic model comparisons or absence of decision-support integration [3]. Furthermore, classical machine-learning techniques such as support vector machines and ensemble classifiers continue to play a valuable role in low-compute or low-data settings but require systematic evaluation to determine their applicability alongside deep architectures in this modern age.

In this paper, we present a comprehensive investigation of both traditional and deep-learning methods for binary waste classification, distinguishing organic from recyclable materials using a dataset of 25 077 images which could be used to build efficient and effective Data-Driven Decision Support System (DSS). Our major contributions are as follows:

1. **Broad model comparison.** We develop and evaluate a custom CNN, three pre-trained architectures (VGG16, ResNet50, InceptionV3) and two lightweight transfer-learning models (DenseNet121, EfficientNetB0), alongside classical classifiers (Random Forest, SVM, AdaBoost), both with and without Principal Component Analysis.

2. **Quantitative insights.** We quantify the performance gains of deep transfer-learning over traditional approaches, demonstrating a 20 pp improvement in accuracy and an ROC-AUC of 0.98 for DenseNet121. We also assess the marginal impact of Principal Component Analysis (PCA) on classical methods.
3. **Decision-support integration.** We propose an architecture for embedding the optimal classification model within a real-time Data-Driven DSS, outlining its potential to enhance resource recovery, reduce landfill dependency and lower lifecycle environmental impacts in smart-city contexts.

*A. Study Aim*

This study aims to propose a Data-Driven DSS using advanced Machine Learning and Deep Learning techniques to enhance the efficiency and accuracy of AI-enabled automated-waste classification, supporting the transition to a circular economy for urban development.

*B. Specific Objectives*

**Develop and Compare ML Models**

Build and evaluate traditional ML models (e.g., Random Forest, SVM. AdaBoost) and CNN-based models for classifying waste into organic and recyclable categories.

**Assess Dimensionality Reduction**

Investigates the impact of PCA on the classification accuracy of traditional ML models.

**Utilise Transfer Learning**

Implement and assess transfer learning from pre-trained models (e.g., VGG16, ResNet50, EfficientNetB0, DenseNet121) to improve waste classification accuracy.

**Optimise Through Hyperparameter Tuning**

Perform hyperparameter tuning to optimise traditional ML models and CNN-based architectures for waste classification tasks.

*C. Research Questions*

Comparative performance

RQ1: How do traditional ML models perform compared with CNN-based models in terms of classification accuracy?

Impact of Dimensionality Reduction

RQ2: How does PCA affect the performance of traditional ML models?

Effectiveness of Transfer Learning

RQ3: How does transfer learning from pre-trained models influence waste classification accuracy?

The remaining parts of of this paper is as follows. Section II explored some related works in this field while Section III details the methodology, preprocessing steps and model architectures. Section IV presents experimental results and comparative analysis. Section IV also discusses potential integration with Data-Driven DSS platforms and implications for circular-economy policy and for smart cities. Conclusively, Section V closes the paper and proposes directions for future research.

## II. RELATED WORK

Automated waste classification has garnered increasing attention as a means to facilitate circular-economy objectives and enhance urban sustainability. Early approaches relied on traditional machine-learning classifiers such as Support Vector Machines (SVM) and decision trees, which require handcrafted feature extraction and struggle with complex visual patterns [3]. For example, Adedeji and Wang [4] demonstrated an SVM-based system achieving moderate accuracy on binary waste datasets, while other researchers explored ensemble methods including Random Forest, and AdaBoost to improve robustness, noting that feature-engineering overhead remains a critical limitation [3].

The advent of deep learning, specifically CNNs, has revolutionised image-based classification tasks. Sheng et al. [6] integrated CNNs with IoT-enabled smart bins, achieving real-time detection and sorting in a pilot smart-city deployment. Li and Chen [7] compared several pre-trained architectures ResNet, VGG, DenseNet and reported that transfer-learning models consistently outperformed custom CNNs, even with limited training data. Further research corroborated these findings, showing DenseNet121 yields superior F1-scores for medical waste categorisation [8]. However, this work was limited to the pre-trained models compared in the study.

Several studies have examined dimensionality-reduction techniques to alleviate computational burdens for traditional classifiers. Jolliffe and Cadima [9] reviewed PCA as an effective method to compress high-dimensional feature spaces, while Ian et. Al. [10] applied PCA prior to SVM and Random Forest, observing marginal improvements and efficiency for dimensionality cutback of high-resolution spatial transcriptomics data. More recent work by Santoso [11] combined PCA with transfer-learning outputs but found that deep feature representations alone often render dimensionality reduction unnecessary.

Decision-support integration represents another vital strand of research. AI and machine-learning-driven DSS in solid-waste management are explored in several studies as reported in this survey [12]. Most of the work concluded that real-time analytics can significantly reduce landfill diversion times and enhance decision-making. Some research also involves implementing an MCDA-based DSS incorporating CNN predictions for optimal routing of waste collection vehicles in urban environments [13]. The study presents an AI-driven waste classification model that incorporates IoT and Blockchain technologies. Research has also demonstrated that embedding classification outputs within a multi-criteria framework enhances stakeholder decision-making for facility location and capacity planning [14], hence, the need for a robust Data-Driven DSS.

Despite these advances, existing literature often lacks a holistic comparison of classical ML, custom CNNs, and multiple transfer-learning models on the same large-scale dataset, as well as a clear pathway for embedding the best model within an operational DSS for smart cities projects. Our work addresses this gap by benchmarking six deep-learning architectures against three traditional classifiers (with and without PCA) on a 25 077-image dataset, and by outlining a deployment architecture tailored to smart-city infrastructures.

### III. METHODOLOGY

This study employs a combination of deep learning, transfer learning, and traditional machine learning techniques to classify waste images into two categories which are Organic Waste and Recyclable Waste for proper waste management. The overview of the work is shown in Figure 2 where the outputs of the waste classification lead to the implementation of the decision support system proposed in [2]. The methodology includes dataset preprocessing, model development, hyperparameter tuning, and model evaluation of the model for a responsible, reliable, efficient and sustainable solution.

*A. Dataset Collection and Specifications*

The dataset used in this study was extracted from Kaggle [15], consisting of 25,077 images categorised into Organic and Recyclable waste classes. For training purposes, the dataset was split into, training (80%) and testing (20%). The training directory consists of 22564 images, while the test directory consists of 2513 images. The dataset was created to support sustainability research and some random images from the dataset are displayed in Figure 1.

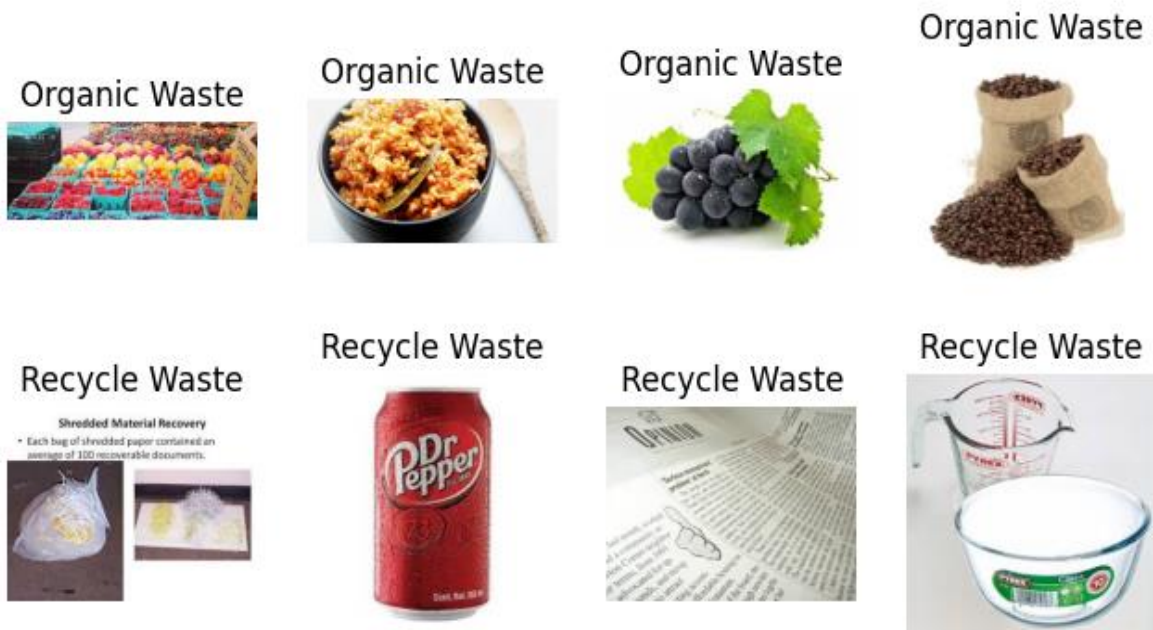

*Figure 1: Random Images from the dataset*

Partly funded by the University of Hull.

*B. Data Enhancement and Visualisation*

To gain insights into data distribution, random Images from each class were visualised as seen in Figure 1, and the label distribution was assessed using pie and bar charts as shown in Figure 3. The images were loaded into arrays using cv2, converted to RGB for uniformity, and resized to a standard dimension of 150 x 150 pixels to maintain consistency in input to the model.

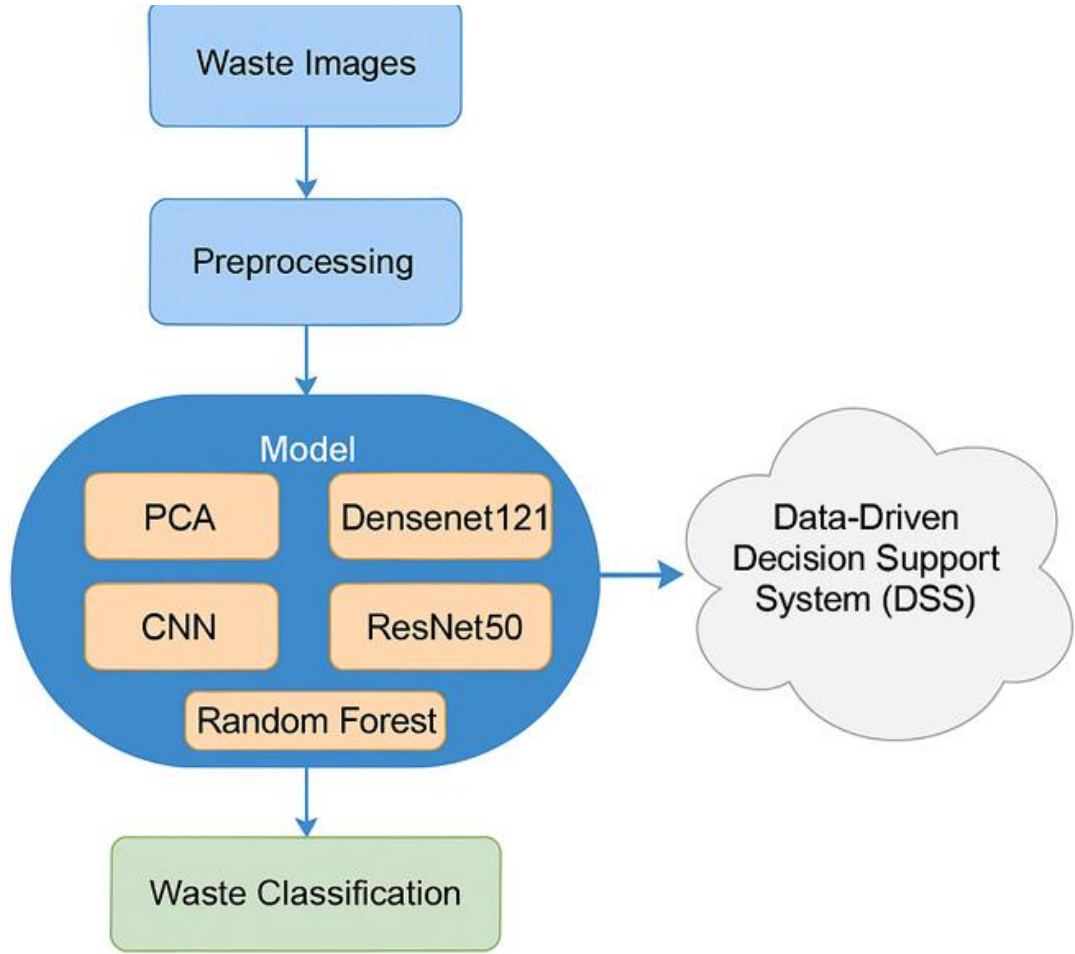

*Figure 2: AI-Enabled Waste Classification as Data-Driven Decision Support Tool.*

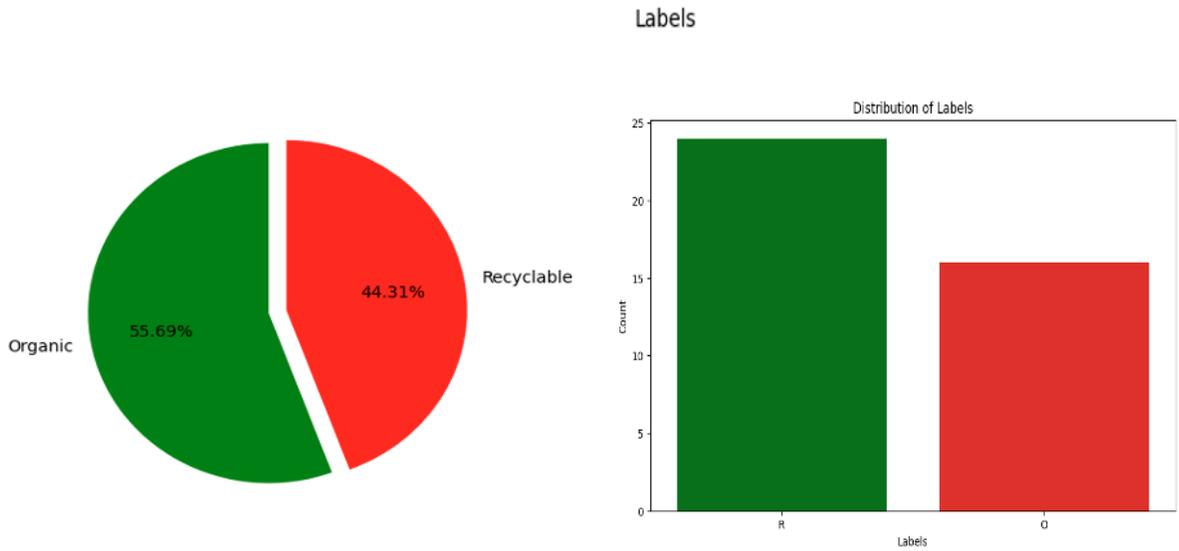

*Figure 3: Images and Labels Distributions*

*C. Data Preprocessing*

Advanced preprocessing techniques were applied to enhance feature extraction, including converting images to grayscale and employing Canny edge detection. The images were normalised by scaling pixel values between 0 and 1. Data augmentation techniques, such as widths and height shifts and horizontal flips were applied using ImageDataGenerator to increase data variability and improve model robustness.

*D. CNN Architecture Model*

A Convolutional Neural Network was designed with sequential layers for the classification tasks. The model architecture includes convolutional layers which consist of filter and kernel with Rectified Linear Unit (ReLU) activation function which introduces non-linearity into the network, max-pooling layers, flatten layer, a fully connected layer that connects every neuron in one layer to every neuron in the next layer, and output layer represents the predicted class. The output layer utilises a sigmoid activation function for the binary classification. Regularisation techniques, including Early Stopping, were employed to prevent overfitting. The model was compiled with the Adam optimizer and binary cross-entropy loss function.

*E. Transfer Learning Models*

Transfer Learning Models were constructed using pretrained models like VGG16, ResNet50, EfficientNetB0, DenseNet121, and InceptionV3. Custom classification layers were added to these models and were fine-tuned to optimise the performance of the waste classification datasets. DenseNet121 (see Figure 4 for its architecture) was particularly effective, as shown in prior studies on medical image classification [16, 17].

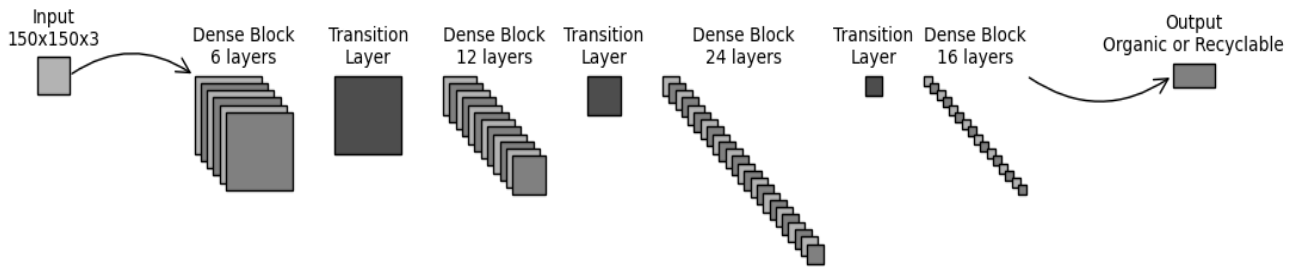

*Figure 4: DenseNet121 Architecture*

This architecture starts with an input layer with a size of 150x150 x3 which is the preprocessing stage where the input data is prepared for the network. Thereafter, the images pass through the series of dense blocks and transition layers after the preprocessing. The dense blocks depicted the layers labelled Dense Block 1 to 4 consist of four different layers 6, 12, 24, and 16 layers respectively. Each layer within the dense blocks receives inputs from previous layers thereby enhancing efficiency and promoting feature reuse. To manage the complexity of the model and reduce the size of feature maps transition layers are placed between dense blocks. Lastly, the output layer which is the last layer labelled organic, and recyclable illustrates the network's tasks which is to classify the images into two groups organic and recyclable waste as shown in Figure 4.

*F. Model Training*

The CNN model was trained on the dataset with 22564 images, splitting the images into training (80%) and testing. (20%) belonging to 2 classes Organic and Recyclable with a batch size of 128 at 0.001 learning rate. Early stopping was applied to monitor the model and prevent overfitting. The pre-trained models were compiled and trained on the custom layers and the pre-trained layers remained unchanged. However, the following models, like Random Forest, XGBoost., Naïve Bayes, AdaBoost, SVM, and Logistic Regression were implemented and trained with reduced data dimensions using PCA and without PCA to enhance computational efficiency and assess the model's performance in predicting circular economy results [3, 18]. Other studies also employed a stacking ensemble classifier combining multiple models to enhance classification performance.

*G. Model Evaluation*

The study used accuracy, a classification report based on a confusion matrix, and ROC-AUC as the evaluation metrics after the training process. The performance of several models will be used to identify the most effective techniques for supporting decision-making in circular economy strategies as demonstrated in other studies [19, 20]. Confusion matrix is widely used as a visual evaluation tool employed in machine learning. It depicts the predicted class results on the columns and the actual class on the rows [21]. Naidu et al. [22] proposed that accuracy, precision, recall, and f1-score are the most used evaluation metrics although might be limited when used in isolation therefore other metrics like the ROC-AUC should be considered because they provide further insight into the effectiveness of the algorithms and very thoughtful when evaluating the effectiveness of binary classification tasks.

## IV. RESULTS

To be able to determine a responsible automated waste classification model for smart cities, the performances of CNN and Transfer Learning models were evaluated using accuracy loss function and other metrics during the training and validation phases. Binary Cross-Entropy was utilised as the loss function, measuring the error between predicted and actual labels, while accuracy measured the proportion of correct predictions. These metrics are standard for evaluating and optimising models in binary classification tasks. Additionally, Precision, Recall, F1-score, and Receiver Operating Characteristics (ROC) / Area Under Curve (AUC) were used to further assess model performance.

However, traditional machine learning models like Random Forest, Naïve Bayes, AdaBoost, XGBoost, SVM, and Logistic Regression were also employed in the study to compare their performance against deep learning models which relates to RQ1. The same metrics - Accuracy, Precision, Recall, and F1-score were used to assess these models.

There is no significant difference in hyperparameter tuning for the traditional machine learning models with and without PCA in terms of accuracy and other metrics such as precision, recall, and f1-score.

The Ensemble model, evaluated with and without PCA, demonstrated how hyperparameter tuning impacts their performance. Stacking models, both tuned and untuned, with or without PCA, showed a consistent accuracy of 70-72%. Although PCA slightly improved some models, but the overall impact was minimal. An overview and comparison of the results is presented in Figure 5.

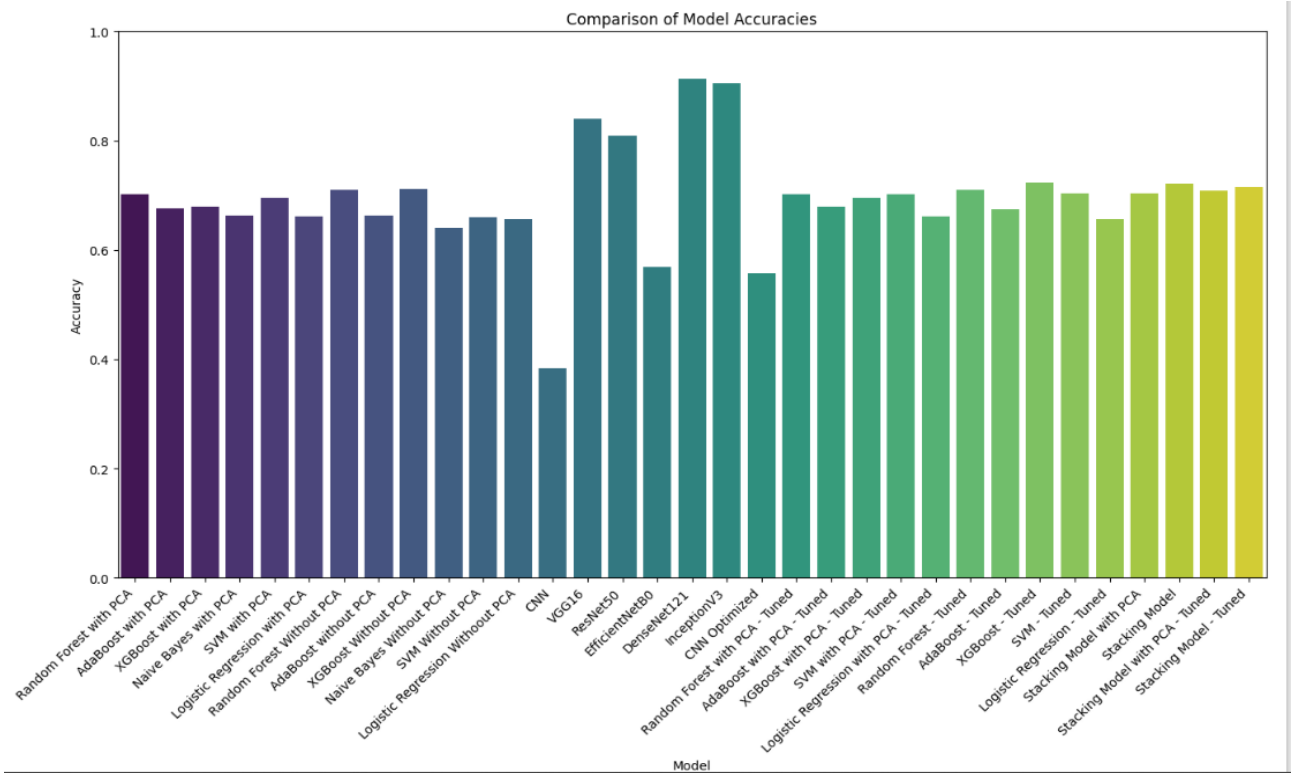

*Figure 5: Comparison of Model Accuracies.*

A. *Receiver Operation Characteristics (ROC) Analysis*

Figure 6 presents ROC-AUC curves, where DenseNet121 displayed the highest AUC of 98%, confirming its superior performance compared to other deep learning models. This high AUC suggests DenseNet121's robustness in distinguishing between the two classes.

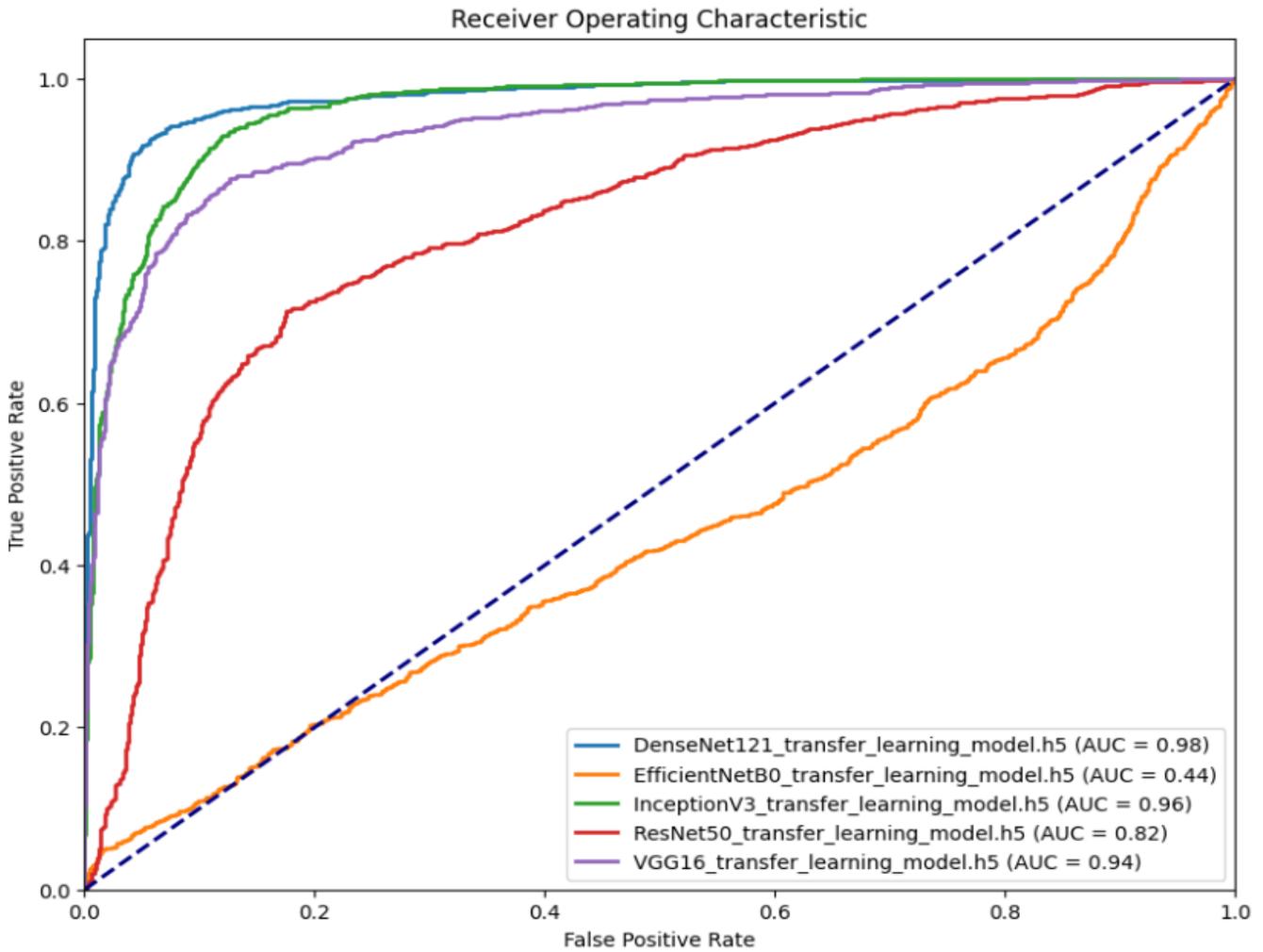

*Figure 6: Receiver Operating Characteristic (ROC).*

### B. DISCUSSION

The study compared deep learning models with traditional machine learning techniques to evaluate their performance in waste classification tasks for smart cities and circular economy principles integration. Additionally, the impact of PCA on traditional machine learning models was analysed to assess its effectiveness in improving classification accuracy and to assess any potential benefits.

*1) Deep Learning Model Performance*

The results answered RQ3 indicating that DenseNet121 achieved the highest overall performance, with an accuracy of 91% in classifying waste into organic and recyclable categories. DenseNet121 also exhibited a balanced performance between precision and recall, with F1-scores of 93% for organic waste and 89% for recyclable waste. The F1-score, which represents the harmonic mean of precision and recall, is a balanced metric that accounts for both performance measures. InceptionV3 also generalised well, though it performed slightly below DenseNet121. VGG16 proved effective but had some difficulty predicting recyclable wastes in class R. Zulhusni et al. [23] proposed a DenseNet model for waste classification and achieved an overall accuracy of 93%.

ResNet50 showed good precision and recall, with an F1-score of 84% for organic waste (class O) and 76% for recyclable waste (class R). While ResNet50 demonstrated balanced performance, it fell slightly below compared to DenseNet121 and InceptionV3. EfficientNetB0 did not generalise well, particularly for recyclable waste, and requires further improvement. The CNN model achieved a balanced performance with 72% accuracy, but it showed significant variation in precision and recall between the two classes. Optimised CNN models, however, did not perform well and were deemed unreliable for class R predictions.

Similarly, Aral et at. [24] confirmed the effectiveness and superiority of DenseNet121 over other transfer learning models they trained. While previous research suggested that models like ResNet50 and EfficientNetB0 achieved higher accuracy in waste classification tasks, this study supports the finding that DenseNet121 leads to more efficient recycling processes [25].

Figure 6, which shows ROC-AUC curves, indicates the True Positive Rate (TPR) and False Positive Rate (FPR), both of which range from 0.0 to 1.0. DenseNet121 demonstrated the highest AUC of 98%, indicating it had the best performance compared to other deep learning models.

*2) Traditional Machine Learning Model Performance (With and Without PCA)*

The results also included the performance of traditional machine learning models such as XGBoost, Naïve Bayes, SVM, and Logistic Regression with and without PCA in classifying organic and recyclable waste. Random Forest and XGBoost with PCA achieved the highest accuracy of 70% and 68%, respectively, with f1-scores of 76% and 73%. Without PCA both models achieved 71% accuracy and F1-scores of 77% and 66%, respectively. These results indicate that both models perform well with and without PCA, maintaining a balance between precision and recall, making them effective for waste classification answering RQ2.

AdaBoost and SVM, both with and without PCA, showed slightly lower accuracy but maintained high precision and recall scores. Naïve Bayes had the lowest accuracy in both scenarios but still maintained balanced F1- scores. The results suggest that the effectiveness of PCA depends on the specific dataset and model. In some cases, models performed better without PCA, indicating that dimensionality reduction may not always be necessary. For models like Random Forest and XGBoost, PCA slightly decreased accuracy and F1-scores, suggesting that the original features may be more suitable without dimensional reduction. Kumar et al. [26] explored SVM classifier for waste classification, and it achieved 96.5% accuracy, 95.3% sensitivity, and 95.9% specificity in classifying waste types in terms of circular manufacturing and showed the ability to manage the COVID related medical waste mixed.

The study's findings align with previous research, showing that advanced deep learning models, particularly DenseNet121, significantly outperform traditional machine learning models, also answering RQ1. DenseNet121 demonstrated balanced performance with high precision, recall, and f1-scores for both organic and recyclable wastes, achieving the highest accuracy of 91%. While InceptionV3 and VGG16 also generalize well, they performed slightly below DenseNet121. Traditional models like Random Forest and XGBoost demonstrated good accuracy and balanced performance with the application of PCA, though they still perform below deep learning models. However, despite the performance of the classification accuracy, the major drawback of the study is that the dataset is limited to two waste categories which raises concerns about the generalisability of the model to more diverse and complex real-world contexts.

*C. Contributions*

Broad Model Comparison: The study provides a comprehensive comparison of a custom CNN, three pre-trained architectures (VGG16, ResNet50, InceptionV3), and two lightweight transfer-learning models (DenseNet121, EfficientNetB0), against traditional classifiers (Random Forest, SVM, AdaBoost) with and without PCA. This addresses a gap in existing literature, which often lacks a holistic comparison of these different approaches on a single, large-scale dataset.

Quantitative Insights: The research quantifies the performance gains of deep transfer learning over traditional methods, showing that DenseNet121 achieved the highest accuracy of 91% and an ROC-AUC of 0.98, outperforming the best traditional classifier by 20 percentage points. It also assesses the marginal impact of PCA on traditional models, finding that its benefit was negligible.

Decision-Support Integration: The paper proposes an architecture for integrating the most effective classification model into a real-time Data-Driven DSS. This highlights a pathway for deploying the model to enhance resource recovery, reduce landfill reliance, and lower environmental impacts in smart-city contexts.

## V. CONCLUSION

This study demonstrates the effectiveness of AI-enabled waste classification as a key enabler of circular economy strategies in smart urban environments. By systematically evaluating deep learning and traditional machine-learning approaches, we show that the DenseNet121 transfer-learning model achieves the highest classification performance, with 91 % accuracy and an AUC of 0.98, for distinguishing organic from recyclable waste. These results affirm the viability of deploying such models within Data-Driven DSS to automate waste sorting and optimise resource recovery.

The integration of robust classification models into urban waste-management infrastructure has the potential to significantly reduce landfill dependency, enhance material reuse, and mitigate the environmental impacts of incineration. By supporting timely and accurate waste segregation, this approach contributes directly to the resilience and sustainability objectives of modern smart cities.

Future work should focus on expanding the classification scope to include a broader range of waste categories and deploying the system across heterogeneous real-world datasets. Moreover, hybrid architectures combining vision-based models with contextual metadata (e.g., geolocation, waste source) may further enhance performance. Close collaboration between researchers, municipalities, waste management providers, and policymakers will be essential to scaling such intelligent systems for maximum impact in the global transition toward resilient, sustainable urban communities.